\documentclass[letterpaper, 10 pt, conference]{ieeeconf}

  \usepackage{pgfplots}
  \pgfplotsset{compat=newest}
  \usetikzlibrary{plotmarks}
  \usetikzlibrary{arrows.meta}
  \usepgfplotslibrary{patchplots}
  \usepackage{grffile}
  \usepackage{amsmath}
  \usepackage{blindtext}
  \usepackage[bottom]{footmisc}
  
 \usepackage{url}

\newlength\fwidth
 \usepackage[linesnumbered,ruled]{algorithm2e}

\usepackage{graphicx}
\usepackage{framed}
\usepackage{subcaption}
\usepackage{indentfirst}
\usepackage[]{algorithm2e}
\usepackage{multirow}

\usepackage{amsmath, amssymb,amsthm}
\usepackage{color}
\usepackage{booktabs}
\usepackage{tabularx}

\usepackage{comment}
\usepackage{vector}
\usepackage{hyperref}
\usepackage{cleveref}

\newcommand{\etal}{\textit{et al}. }
\crefname{figure}{Fig.}{Fig.}
\Crefname{figure}{Fig.}{Fig.}
\IEEEoverridecommandlockouts                              

\title{Human Motion Prediction using Semi-adaptable Neural Networks}
\author{Yujiao Cheng*, Weiye Zhao*, Changliu Liu, and Masayoshi Tomizuka\thanks{* Equal contribution.}
\thanks{Y. Cheng and M. Tomizuka are with the Department of Mechanical Engineering, University of California, Berkeley, CA 94720 USA (e-mail: \tt\small yujiaocheng, tomizuka@berkeley.edu).
}\thanks{W. Zhao and C. Liu are with the Robotics Institute, Carnegie Mellon University, Pittsburgh, PA 15213 USA (e-mail: \tt\small weiyez, cliu6@andrew.cmu.edu).}}

\begin{document}

\maketitle

\begin{abstract}
Human motion prediction is an important component to facilitate human robot interaction. Robots need to accurately predict human's future movement in order to safely plan its own motion trajectories and efficiently collaborate with humans. Many recent approaches predict human's movement using deep learning methods, such as recurrent neural networks. However, existing methods lack the ability to adapt to time-varying human behaviors, and many of them do not quantify uncertainties in the prediction. This paper proposes an approach that uses a semi-adaptable neural network for human motion prediction, and provides uncertainty bounds of the predictions in real time. In particular, a neural network is trained offline to represent the human motion transition model, and then recursive least square parameter adaptation algorithm (RLS-PAA) is adopted for online parameter adaptation of the neural network and for uncertainty estimation. Experiments on several human motion datasets verify that the proposed method significantly outperforms the state-of-the-art approach in terms of prediction accuracy and computation efficiency.
\end{abstract}

\section{Introduction}
Smooth interactions among intelligent entities depend on a clear understanding of what the others would do in various circumstances. For example, soccer players predict the motions of their teammates for better cooperation; pedestrians have a notion of where others are going so as to avoid collisions.
Similarly, robots that interact in proximity with humans are required to know what the human is going to do in the near future.
The benefits are that, based on the predictions, robot can plan collision-free trajectories to assure human's safety and schedule their actions in advance to improve task efficiency. 
Human motion prediction has applications well beyond human robot interaction~\cite{yamazaki2012home}~\cite{knight2013smart}~\cite{diftler2011robonaut}, it also plays a key role in computer vision. Adequate prediction of human motion can facilitate 3D people recognition and tracking~\cite{gupta20143d}, motion generation in computer graphics (CG)~\cite{kovar2008motion}, and psychology biological motion modeling~\cite{troje2002decomposing}.

However, human motion is inherently difficult to predict due to the nonlinearity and stochasticity in the human behavior~\cite{peng2014assessing}. 
In addition, individual differences are also prominent. Prediction models that work for one person may not be applicable to another.

Early attempts have been made to predict human motion using kalman filter and particle filter~\cite{kohler1997using}~\cite{bruce2004better}, where the problem is posed as a tracking problem. Another category of approaches assumes that human is rational with respect to certain cost functions. Human trajectories can then be predicted by optimizing the cost function~\cite{kalakrishnan2011stomp}. The difficulty of this method is that the cost functions of human are hard to obtain due to stochasticity and complexity in human intention. Another domain of work prominently focuses on latent variable based probabilistic models. Wu \etal~\cite{wu2014leveraging} use hidden markov models (HMMs) combined with multi-layer perceptrons to model the evolution patterns of motion trajectory.  

Similar to HMMs, recurrent neural networks (RNNs) have distributed hidden states to store information about the past, and many works on RNNs have obtained big success on human motion prediction~\cite{ghosh2017learning}~\cite{alahi2016social}, but they still suffer from several problems. First problem is that RNNs are hard to train. Heroic efforts of many years still fail to accelerate the training speed of RNNs. Second problem is that predictions from RNNs are deterministic, which is not satisfactory in human robot interaction, since the robot needs the uncertainty level of human's future motion for safe motion planning. The last serious problem is that the RNN models are fixed and they cannot adapt to time-varying human behaviors.  

We aim to solve these problems by proposing a semi-adaptable neural network. To be specific, a neural network is trained offline to represent the human motion transition model, and then recursive least square parameter adaptation algorithm (RLS-PAA) is adopted for online parameter adaptation of the last layer in the neural network and for uncertainty estimation. The proposed method advantages human motion prediction in three aspects. First, it is computationally more efficient to use a feedforward neural network than to use a RNN for approximation of the human transition model. In the meanwhile, the mechanism for adaptable feedforward neural networks is equally applicable to adaptable RNNs. Second, it adapts the model to time-varying behaviors and individual differences in human motion, which yields more accurate predictions. Third, it computes the uncertainty level of the predictions, which is important for safe motion planning of robots. To verify the the effectiveness of our adaptation scheme, we compare our method with the state-of-the-art online learning algorithm called the identifier-based algorithm~\cite{ravichandar2017human}. Identifier-based algorithm adapts all the parameters in the offline-trained neural network model online, using gradient descent to minimize the prediction error. Results demonstrate that our method achieves a higher prediction accuracy, and the performance is maintained across a variety of motion categories and motion datasets. Our code is publicly available at \emph{github.com/msc-berkeley/Human-Motion-Prediction}.

The remainder of the paper is organized as follows. \Cref{sec: problem} formulates the human motion prediction problem. \Cref{sec: method} proposes the method of semi-adaptable neural networks. \Cref{sec: result} shows the performance of the proposed method and compares it with other methods. \Cref{sec: discussion} discusses analyses and extensions of the proposed method. \Cref{sec: conclusion} concludes the paper.

\section{Problem Formulation\label{sec: problem}}
Predicting human motion is important for smooth human robot interaction, because first, if the robot knows what the human is going to do, it can adapt its actions to collaborate
with humans in an efficient way, and second, plan collision-free
trajectories to guarantee human’s safety~\cite{rcim2018robot}. This paper concerns the prediction of one human joint (e.g., wrist), which is reasonable because when a human works in close proximity to a robot, special attention should be paid to the movement of human's hand. Moreover, one joint motion prediction
is extendable to that of multiple joints, which will be discussed in \cref{sec: discussion}.

The transition model of human joint motion is formulated as
\begin{equation}
{\mathbf{x}(k+1)}  = f^*(\mathbf{x}^*(k), a) + w_k,
\label{eq: dynamic model}
\end{equation}
where $\mathbf{x}(k+1)\in \mathbb{R}^{3M}$ denotes human's $M$-step positions of the joint at future time steps $k+1, k+2, \ldots, k+M$ in a Cartesian coordinate system. $M\in \mathbb{N}$ is the prediction horizon. Denoting the Cartesian position of the joint at time step $k$ by $p(k)\in\mathbb{R}^3$, $\mathbf{x}(k+1)$ is obtained by stacking $p(k+1), p(k+2), ..., p(k+M)$.
$\mathbf{x}^*(k) \in \mathbb{R}^{3N}$ denotes human's past $N$-step positions of the joint. It is also constructed by stacking the position vectors $p(k), p(k-1), ..., p(k-N+1)$. $a \in \mathbb{N}$ is an action label to distinguish different motions, and this label is obtained by the action recognition module of the system \cite{rcim2018robot}. $w_k \in \mathbb{R}^{3M}$ is a zero-mean white Gaussian noise. 
The function $f^*(\mathbf{x}^*(k), a): \mathbb{R}^{3N} \times \mathbb{N}^1 \to \mathbb{R}^{3M}$ represents the transition of the human motion, which takes historical trajectory and current action label as inputs, and outputs the future positions of the joint. 

Since human behavior differs greatly across individuals and is highly time-varying, function $f^*$ may not be a time invariant function. Though $f^*$ takes the discrete parameter action label $a$ as one of the inputs to accommodate some of the variances, an adaptable model of $f^*$ is still desired to account for continuous changes online in order to provide accurate prediction.

\section{Semi-daptable Neural Networks}\label{sec: method}
Since human's motion is not only time-varying but also highly nonlinear, we propose to use a neural network to construct the model $f^*$, bacause neural networks have good model capacity.
To make it adaptable, notice that if we remove the last layer, the pre-trained neural network becomes an effective feature extractor \cite{athiwaratkun2015feature}, the features from which are better than handcrafted ones~\cite{sharif2014cnn}.
Therefore we only adapt the weights of output layer of the neural network online, which fixes the weights of the remaining layers, hence fixing the extracted features. 

The proposed procedure is that \begin{enumerate}
    \item We first design a neural network architecture;
    \item We train the model $f^*$ offline;
    \item During online execution, we adapt the parameters of the last layer of the neural network using efficient adaptation algorithm;
    \item We then compute the uncertainty level of predictions given the adaptation result.
\end{enumerate}

The algorithm is shown in \Cref{ANN}.
\begin{algorithm}
    \SetKwInOut{Input}{Input}
    \SetKwInOut{Output}{Output}
    \SetKwInOut{Variables}{Variables}
    \SetKwInOut{Initialization}{Initialization}
    \Input{Offline trained neural network \eqref{eq: T1 NN} with $g$, $U$ and $W$}
    \Output{future trajectory $\mathbf{x}(k+1)$}
    \Variables{Adaptation gain $F$, \textit{a priori} mean squared estimation error of states $ X_{\tilde{x}\tilde{x}}$, mean squared estimation error of the parameters $X_{\tilde{\theta}\tilde{\theta}}$, neural network last layer parameters $\theta$, estimated rate of change $\delta\theta$ (approximation of $\Delta\theta$), variance of zero-mean white Gaussian noise $Var(w_k)$}
    \Initialization{$F=1000\mathcal{I}$,
    $\theta=$ column stack of $W$,$ X_{\tilde{x}\tilde{x}}=\textbf{0}, X_{\tilde{\theta}\tilde{\theta}}=\textbf{0}$, $\lambda_1=0.998$, $\lambda_2=1$}
    \While{True}{
    Wait for a new joint position $p$ captured by Kinect and current action label $a$ from action recognition module\;
    Construct $s_k=[p(k), p(k-1), ... , p(k-N+1), a,1]^T$\;
    Obtain $\Phi(k)$ by diagonal concatenation of $max(0, g(U,s_k))$\;
    Update $F$ by (\ref{eq:update_F})\;
    Adapt the parameters $\theta$ in last layer of neural network by (\ref{eq:update_theta})\;
    Calculate future joint trajectory $x(k+1)$ by (\ref{eq:Transformed LTV})\;
    Update $\delta\theta$ and calculate $X_{\tilde{x}\tilde{x}}$ and $ X_{\tilde{\theta}\tilde{\theta}}$ by (\ref{eq:update_x}) and (\ref{eq:parameter-estimation-error-covariance})\;
    send $x(k+1)$ and $X_{\tilde{x}\tilde{x}}$ to robot control. 
    }
    \caption{Semi-adaptable neural network for human motion prediction}
    \label{ANN}
\end{algorithm}

\subsection{Training the Neural Network}
To train the transition model $f^*$, we choose an $n$-layer neural network with ReLU activation function which takes the positive part of the input to a neuron.
\begin{align}
f^*(\mathbf{x}^*(k), a)  = W^T \max(0, g(U, s_k)) + \epsilon(s_k),\label{eq: T1 NN}
\end{align}
where $s_k = [\mathbf{x}^*(k)^T, a,1]^T \in \mathbb{R}^{3N+2}$ is the input vector, $g$ denotes $(n-1)$ - layer neural network, whose weights are packed in $U$. $\epsilon(s_k) \in \mathbb{R}^{3M}$ is the function reconstruction error, which goes to zero when the neural network is fully trained. $W\in\mathbb{R}^{n_h\times 3M}$ is the weights of the last layer, where $n_h \in \mathbb{N} $ is the number of neurons in the hidden layer of the neural network~\cite{ravichandar2017human}.


\subsection{Parameter Adaptation Algorithm}
To accommodate both the time varying behavior of human and individual differences among different people, it is important to update the parameters online. In this paper, we applied the recursive least square parameter adaptation algorithm (RLS-PAA) with forgetting factor~\cite{goodwin2014adaptive} to asymptotically adapt the parameters in the neural network. 

By stacking all the column vectors of $W$, we get a time varying vector  $\theta\in\mathbb{R}^{3Mn_h}$ to represent the weights of last layer. $\theta_k$ denotes its value at time step $k$. To represent the extracted features, we define a new data matrix $\Phi_k \in\mathbb{R}^{3M\times 3Mn_h}$ as a diagonal concatenation of $M$ pieces of $\max(0, g(U,s_k))$. Using $\Phi_k$ and $\theta_k$, \eqref{eq: dynamic model} and \eqref{eq: T1 NN} can be written as
\begin{equation}
	\mathbf{x}(k+1) =\Phi_k \theta_k+w_k.\label{eq:Transformed LTV}
\end{equation}
Let $\hat{\theta}_k$ denotes the parameter estimate at time step $k$, and let $\tilde{\theta}_k=\theta_k-\hat{\theta}_k$ be the parameter estimation error. We define the \textit{a priori} estimate of the state and the estimation error as:

\begin{align}
	\hat{\mathbf{x}}\left(k+1|k\right) =&\Phi_k\hat{\theta}_k,\label{eq:prediction}\\
	\tilde{\mathbf{x}}\left(k+1|k\right) =&\Phi_k\tilde{\theta}_k+w_k.
\end{align}

The core idea of RLS-PAA is to iteratively update the parameter estimation $\hat{\theta}_k$ and predict  $\mathbf{x}(k+1)$ when new measurements become available. The parameter update rule of RLS-PAA can be summarized as:
\begin{equation}
	\hat{\theta}_{k+1}=\hat{\theta}_k+F_k\Phi^{T}_k\tilde{\mathbf{x}}\left(k+1|k\right),
\label{eq:update_theta}
\end{equation}
where $F_k$ is the learning gain updated by:
\begin{equation}
\begin{split}
&F_{k+1}\\
&= \frac{1}{\lambda_1(k)}[F_k - \lambda_2(k)\frac{F_k\Phi_k\Phi^T_kF_k}{\lambda_1(k)+\lambda_2(k)\Phi^T_kF_k\Phi_k}]
\end{split}\label{eq:update_F}
\end{equation}
where $0<\lambda_1(k)\leq1$ and $0<\lambda_2(k)\leq2$.
Typical choices for $\lambda_1(k)$ and $\lambda_2(k)$ are:
\begin{enumerate}
  \item $\lambda_1(k) = 1$ and $\lambda_2(k) = 1$ for standard typical least squares gain.
  \item $0 < \lambda_1(k) < 1$ and $\lambda_2(k) = 1$ for least squares gain with forgetting factor.
   \item $\lambda_1(k) = 1$ and $\lambda_2(k) = 0$ for constant adaptation gain.
\end{enumerate}

\subsection{Mean Squared Estimation Error Propagation}
To guarantee safety, the uncertainty of the prediction is also quantified during online adaptation ~\cite{liu2015safe}.

\paragraph{State estimation}
Note that $\hat{\theta}_k$ only contains information up to the $\left(k-1\right)$th time step, and $\tilde{\theta}_k$
is independent of $w_k$. Thus the \textit{a priori} mean squared estimation error (MSEE) $X_{\tilde{x}\tilde{x}}\left(k+1|k\right)=E\left[\tilde{\mathbf{x}}\left(k+1|k\right)\tilde{\mathbf{x}}\left(k+1|k\right)^{T}\right]$
is
\begin{equation}
	X_{\tilde{x}\tilde{x}}\left(k+1|k\right)=\Phi_kX_{\tilde{\theta}\tilde{\theta}}(k)\Phi^{T}_k+Var(w_k),\label{eq:update_x}
\end{equation}
where $X_{\tilde{\theta}\tilde{\theta}}(k)=E\left[\tilde{\theta}_k\tilde{\theta}_k^{T}\right]$
is the mean squared error of the parameter estimate and $Var(w_k)$ is the variance of zero-mean white Gaussian noise.

\paragraph{Parameter estimation}
Since the system is time varying, $\Delta\theta_k=\theta_{k+1}-\theta_k\neq 0$. According to parameter estimation algorithm in \eqref{eq:update_theta}, the parameter estimation error is
\begin{equation}
	\tilde{\theta}_{k+1}=\tilde{\theta}_k-F_k\Phi^{T}_k\tilde{\mathbf{x}}\left(k+1|k\right)+\Delta\theta_k\label{eq: belief space paa}.
\end{equation}
The estimated parameter is biased and the expectation of the error can be expressed as
\begin{align}
	E\left(\tilde{\theta}_{k+1}\right)= &\left[I-F_k\Phi^{T}_k\Phi_k\right]E\left(\tilde{\theta}_k\right)+\Delta\theta_k\nonumber \\
	= & \sum_{n=0}^{k}\prod_{i=n+1}^{k}\left[I-F_i\Phi^{T}\left(i\right)\Phi\left(i\right)\right]\Delta\theta_n\label{eq:expectation of parameter estimation error}.
\end{align}

The mean squared error of parameter estimate follows from (\ref{eq: belief space paa}) and (\ref{eq:expectation of parameter estimation error}):
\begin{align}
	& X_{\tilde{\theta}\tilde{\theta}}\left(k+1\right) \nonumber\\
	= & F_k\Phi^{T}_kX_{\tilde{x}\tilde{x}}\left(k+1|k\right)\Phi_kF_k\nonumber  -X_{\tilde{\theta}\tilde{\theta}}(k)\Phi^{T}_k\Phi_kF_k \\
	&-F_k\Phi^{T}_k\Phi_kX_{\tilde{\theta}\tilde{\theta}}(k)\nonumber 
	 +E\left[\tilde{\theta}_{k+1}\right]\Delta\theta^{T}_k\\
	 &+\Delta\theta_kE\left[\tilde{\theta}_{k+1}\right]^{T} -\Delta\theta_k\Delta\theta_k^{T}+X_{\tilde{\theta}\tilde{\theta}}(k).\label{eq:parameter-estimation-error-covariance}
\end{align}

Since $\Delta\theta_k$ is unknown in (\ref{eq:expectation of parameter estimation error}) and (\ref{eq:parameter-estimation-error-covariance}), we define $d\theta_k = \hat{\theta}_k - \hat{\theta}_{k-1}$, and approximate $\Delta\theta_k$ as $\delta \theta_k$ which is the average of $d\theta_i, i = k-n_w+1, k-n_w,...,k$, where $n_w\in\mathcal{N}$ is the window size. 

At step $k$, the predicted trajectory $\hat{\mathbf{x}}(k+1|k)$ together with the uncertainty matrix $X_{\tilde{x}\tilde{x}}(k+1|k)$ is then sent to robot control to generate the safety constraint.
\section{Results\label{sec: result}}

\subsection{Human Motion Prediction on Kinect Data}

\subsubsection{Data Acquisition}
\begin{figure}[b]
\begin{center}
\includegraphics[height = 0.7\linewidth, width=.7\linewidth]{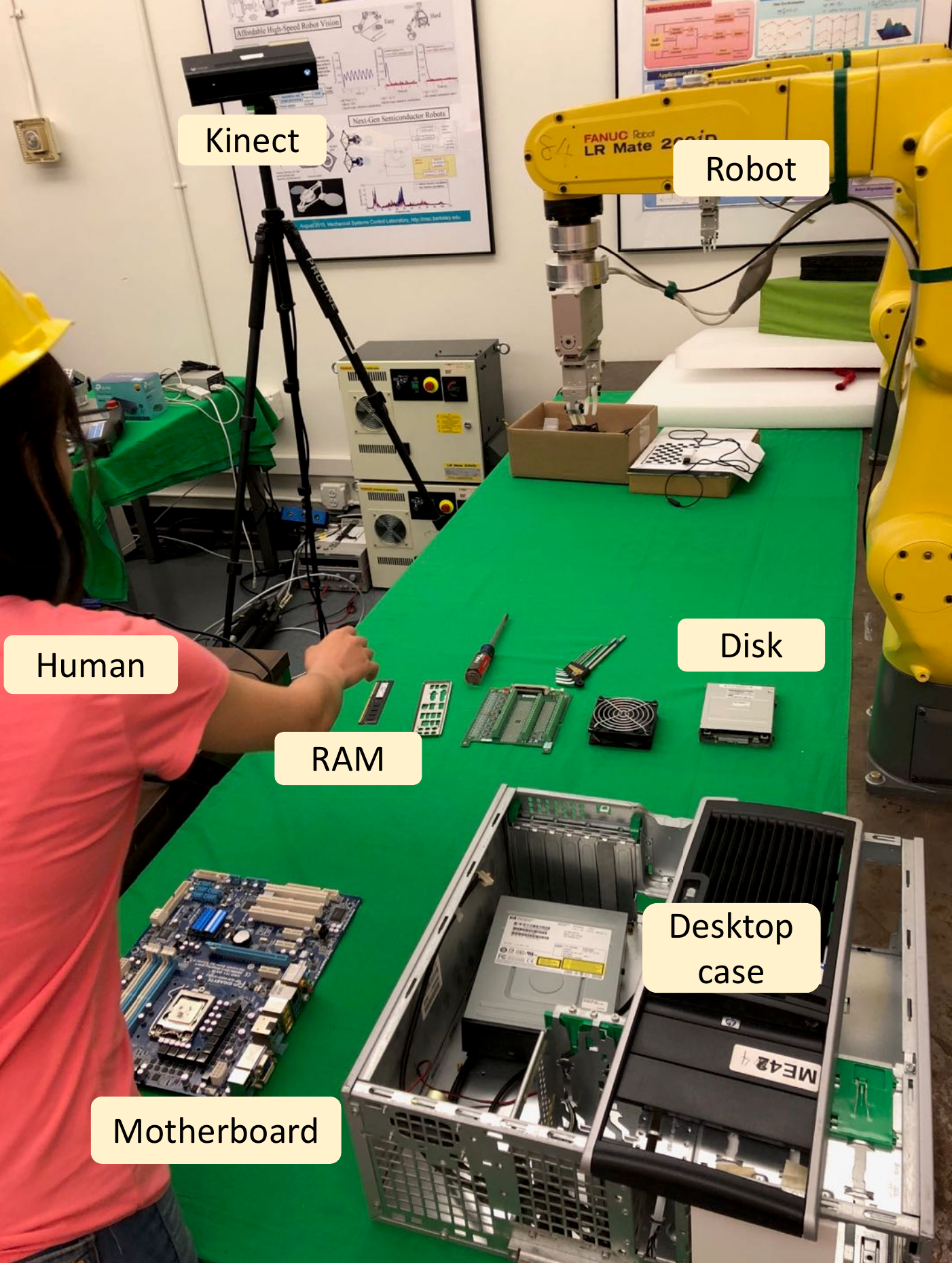}
\caption{Experiment Setup.}
\label{fig: setup}
\end{center}
\end{figure}

In order to verify the proposed human motion prediction approach, experiments are conducted when a human and a robot collaborate to assemble a desktop. The experimental setup is shown in \cref{fig: setup}. A Kinect for Windows v2 is utilized to capture the trajectory of human right wrist at an approximate frequency of 20$Hz$. Human has two options: one is to obtain and insert the RAMs in the motherboard, and the other is to fetch and assemble the disk to the desktop case. $50$ trajectories for each motion are obtained, of which $80\%$ is utilized for offline neural network training, and the remaining is for the validation of the online adaptation algorithm. To smooth the trajectories, we use a low-pass filter $p_s(k) = 0.6\hat{p}(k-1) + 0.4\hat{p}(k)$. $p_s(k)\in \mathbb{R}^3$ is the smoothed position of the joint at time step $k$, which is the weighted average of joint positions $\hat{p}(k-1)$ and $\hat{p}(k)$ measured at time step $k-1$ and $k$. We set the number of past and future joint positions $N$ and $M$ both to be $3$, which implies that we are doing predictions of three time steps approximately $0.15$s. The prediction horizon can be controlled by adjusting the magnitude of $M$ according to the specific applications.

\subsubsection{Neural Network Training}
We use a 3-layer neural network with $40$ nodes in the hidden layer.  The number of nodes in the input layer and the output layer is $9$. The loss function is set to be L2 loss. The learning rate is set to $0.001$ and the number of epochs is $100$. 
\subsubsection{Online Adaptation}
\begin{figure}
\setlength\fwidth{0.3\textwidth}
\pgfplotsset{every axis/.append style={
		 style = {font=\footnotesize}
		 }}
\centering
\input{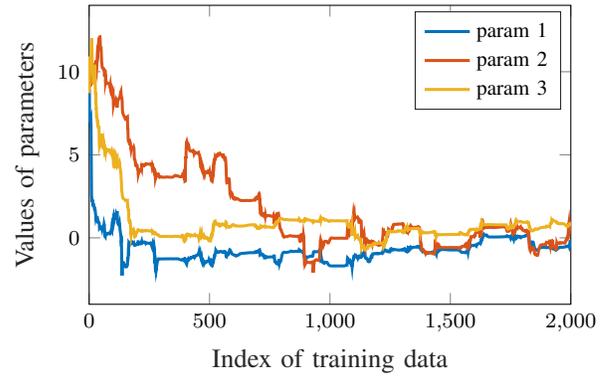}
\caption{Parameter adaptation results.}
\label{fig: converge}
\end{figure}

After obtaining the neural network model for motion transition, we use RLS-PAA to adapt the weights of the last layer. Since the number of nodes in the last layer is $9$, and each node has $41$ parameters, of which $40$ correspond to the outputs from the hidden layer, and the remaining one parameter is a bias term. In total, there are $369$ parameters to be adapted online. In the case that the initial values for all the parameters are set to be $10$, the three parameters we randomly choose quickly settle into finite bounds as shown in \cref{fig: converge}. 

\subsubsection{Trajectory Prediction}
\begin{figure*}
     \centering
    \begin{subfigure}[t]{0.32\textwidth}
        \raisebox{-\height}{\includegraphics[width=\textwidth]{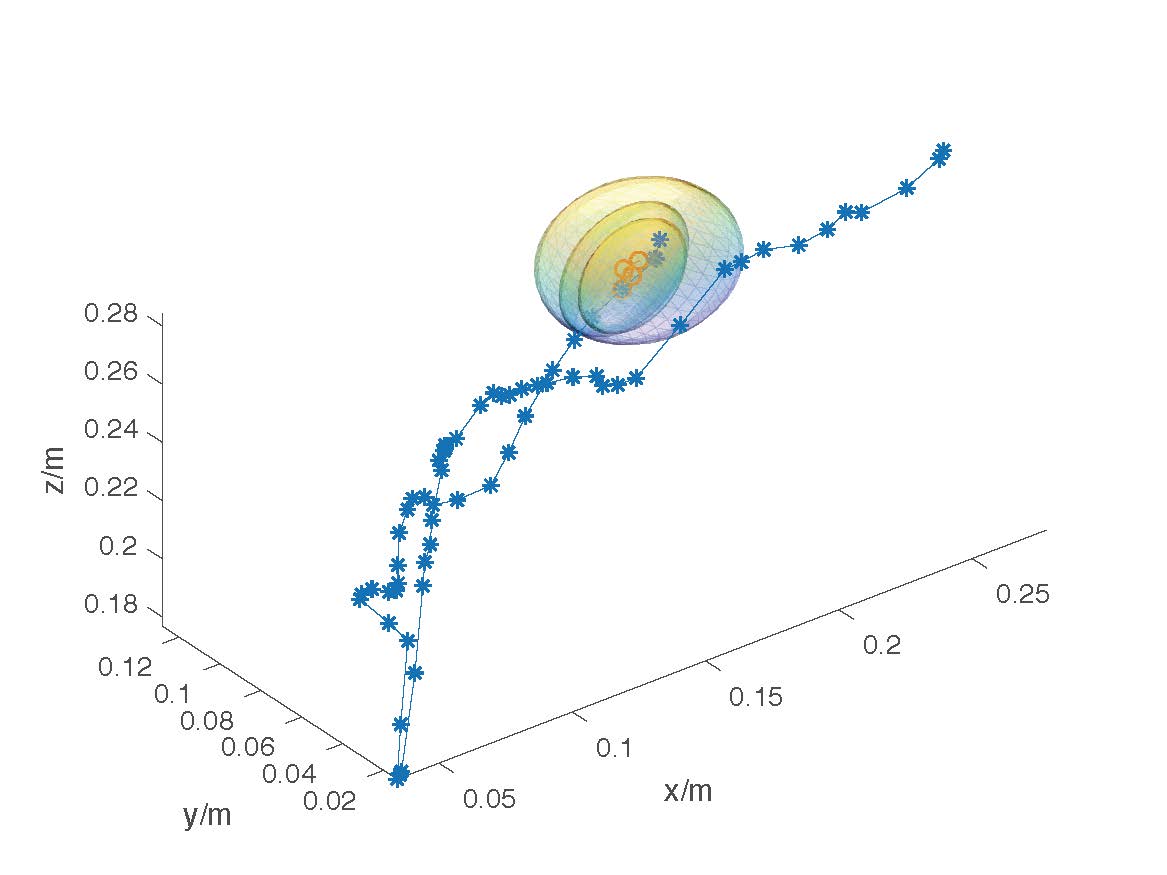}}
        
    \end{subfigure}
    \hfill
    \begin{subfigure}[t]{0.32\textwidth}
        \raisebox{-\height}{\includegraphics[width=\textwidth]{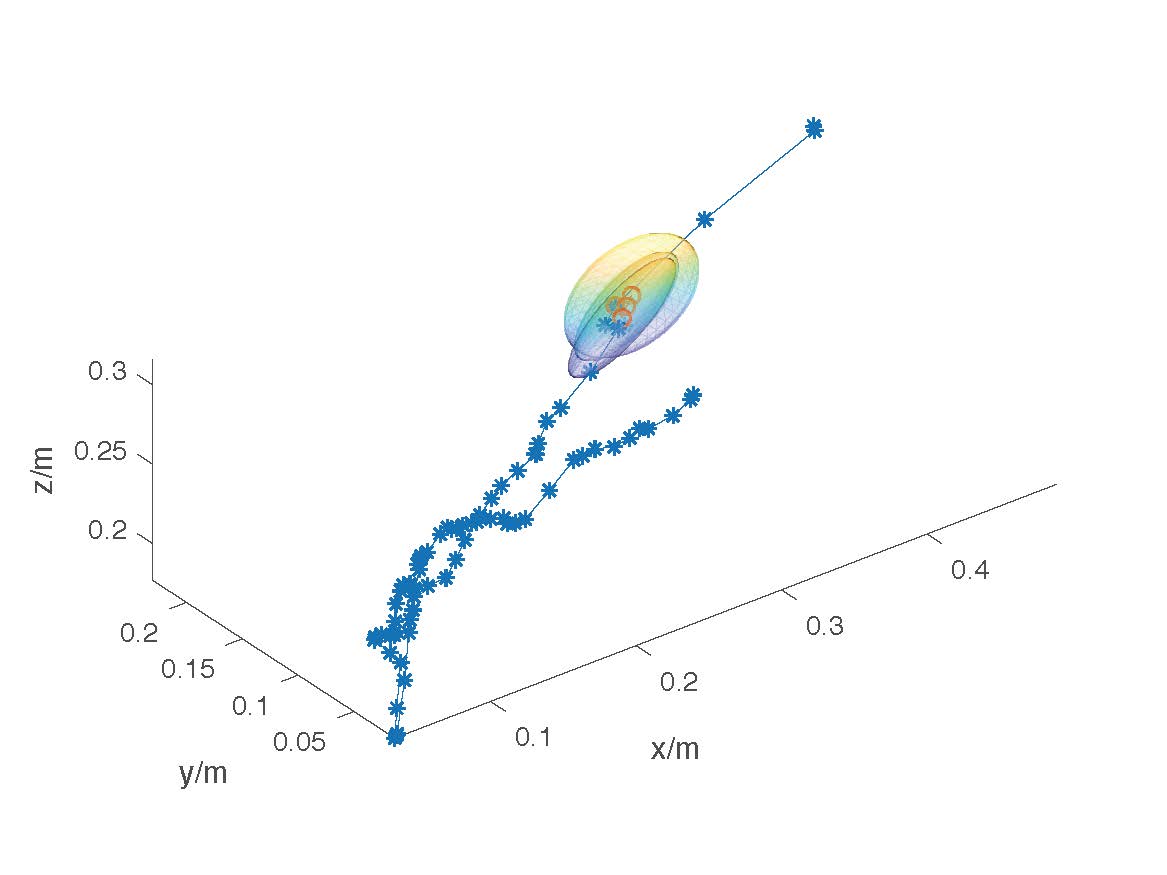}}
      
    \end{subfigure}
    \hfill
    \begin{subfigure}[t]{0.32\textwidth}
        \raisebox{-\height}{\includegraphics[width=\textwidth]{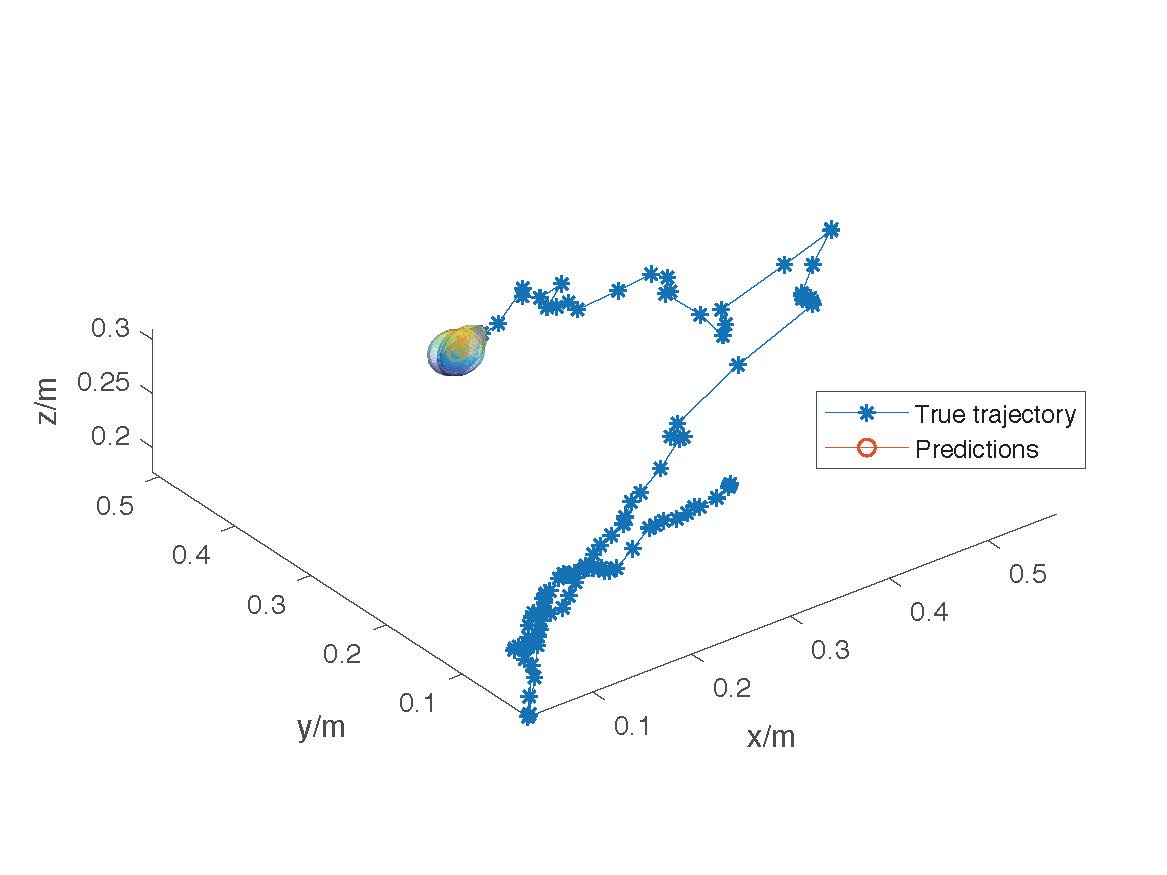}}
       
    \end{subfigure}
    \caption{Joint trajectory prediction with $5\%$ error ellipsoids. Red circles are three predictions of the human wrist, and blue stars are the historical joint trajectory and the three future points. The three subplots represent three moments in human motion of fetching and assembling disk to the desktop case. In the first two subplots, human is bringing the disk near the desktop case and in the third subplot, human is inserting the disk}
 \label{traj}
\end{figure*}

\Cref{traj} shows the prediction results of the right wrist. The three subplots represent three moments in human motion of fetching and assembling disk to the desktop case. In the first two subplots, human is bringing the disk near the desktop case and in the third subplot, human is inserting the disk. Suppose that the current time step is $k$, the blue line demonstrates the true trajectory from the beginning to $k+3$ time steps, of which $3$ ground truth future points are used to compare with predictions. Prediction points are denoted by red circles, and the ellipsoids around them are $5\%$ error bounds, which means that we are $95\%$ sure that the actual motion will be in the ellipsoids. Generally, the performance is good in that the prediction points are near the ground truth, and the $5\%$ error ellipsoids bound the ground truth in most situations. Still, prediction fails sometime when motion of human changes too rapidly as shown in the middle of \cref{traj}. The joint suddenly moves rapidly, and the prediction cannot track this change, so the performance deteriorates.

\subsection{Artificial Systems}

Human motion data obtained from Kinect inherently has measurement noise.
Note that several factors can contribute to the effectiveness of human motion prediction, including noise-reduced measured data and well-defined adaptation procedure. In order to gain a deeper insight into the mechanism behind the proposed algorithm, we artificially simulate the human motion, where the scale of measurement noise and the noise-free trajectories can be controlled. We construct two types of artificial systems, the time varying system and the time invariant system, to test our proposed algorithm. The performance of the proposed algorithm is also compared to identifier-based algorithm on those artificial systems.
\subsubsection{Time Varying System (TV)}
In human dynamics model \eqref{eq: dynamic model}, the future trajectory depends on the past trajectory $\mathbf{x}^*$. However, the trajectory itself is indeed a 3D curve parameterized by a one dimensional parameter. Here we parameterize the trajectory by time $t$. We use a polynomial function to represent the $(x,y,z)$ trajectory: 
\begin{align*}
x(t) = a_x {(t+w)}^2 + b_x (t+w)\\
y(t) = a_y {(t+w)}^2 + b_y (t+w)\\
z(t) = a_z {(t+w)}^2 + b_z (t+w)
\end{align*} 
where $w$ denotes the artificial noise and $t\in[0,5s]$. The trajectory is sampled every $0.05s$. It is easy to verify that given the 3D curve, the resulting dynamic model in the form of \eqref{eq: dynamic model} is time-varying. We have applied two sets of parameters $[a_x, b_x, a_y, b_y, a_z, b_z]$ to generate two different simulated motions, $[0.4, -2, 0, 0.9, 0, 1.05]$ and  $[0.41, -1.9, 1, 0.9, 0, 0.95]$, the unit for three axes is $m$. We also add artificial noise to the artificial motion trajectory data since measurement noise exists in real world data. Therefore, uniformly distributed noise $w$ in the range $[-1,1]$ is added to $t$. For each motion kind we simulate 50 independent trials. 

\subsubsection{Time Invariant System (TI)}
In contrast to the time varying system, the time invariant system has constant model $f$ that does not depend on time. Here we also choose a quadratic formula to parameterize the time-invariant state transition model:
\begin{align*}
x(t + \Delta t) = a_x x(t)^2 + b_x x(t) + w\\
y(t + \Delta t) = a_y y(t)^2 + b_y y(t) + w\\
z(t + \Delta t) = a_z z(t)^2 + b_z z(t) + w.
 \end{align*}
We obtain the TI trajectory with the following strategy: we first randomize the initial position, and then at each sample iteration we get a new position using the quadratic state transition model above provided with the position from last iteration. In the experiment setting, we use two sets of parameters $[a_x, b_x, a_y, b_y, a_z, b_z]$ to represent two kinds of motion classes as well, $[0.06, 0.92, 0, 0.9, 0, 1.05]$ and $[0.061, 0.93, 0, 1.05, 0, 0.96]$. The unit for three axes is $dm$ and the sample time $\Delta t = 0.05s$. Uniformly distributed noise is also added to the simulated data.
\subsection{Comparison} \label{sec: comparison}
To demonstrate the effectiveness of the proposed method, we evaluate the proposed algorithm on four different human motion datasets. They are artificial time varying motion data, artificial time invariant motion data, human motion data collected by Kinect and CMU motion dataset\footnote{Available at
\url{http://mocap.cs.cmu.edu/motcat.php}}. From CMU dataset, we choose to use Mocap data of walking, runing and jumping of all subjects.
To verify our algorithm, we compare it with the state-of-the-art online learning algorithm called Identifier-based algorithm. Identifier-based algorithm adapts all the parameters in the offline-trained neural network model online using gradient descent to minimize the prediction error. Same architecture of the neural network as specified in~\cite{ravichandar2017human} is utilized for fair comparison. Both neural networks are trained offline for $100$ epochs before online adaptation. 

\begin{figure*}
     \centering
    \begin{subfigure}[t]{0.32\textwidth}
        \raisebox{-\height}{\includegraphics[width=\textwidth]{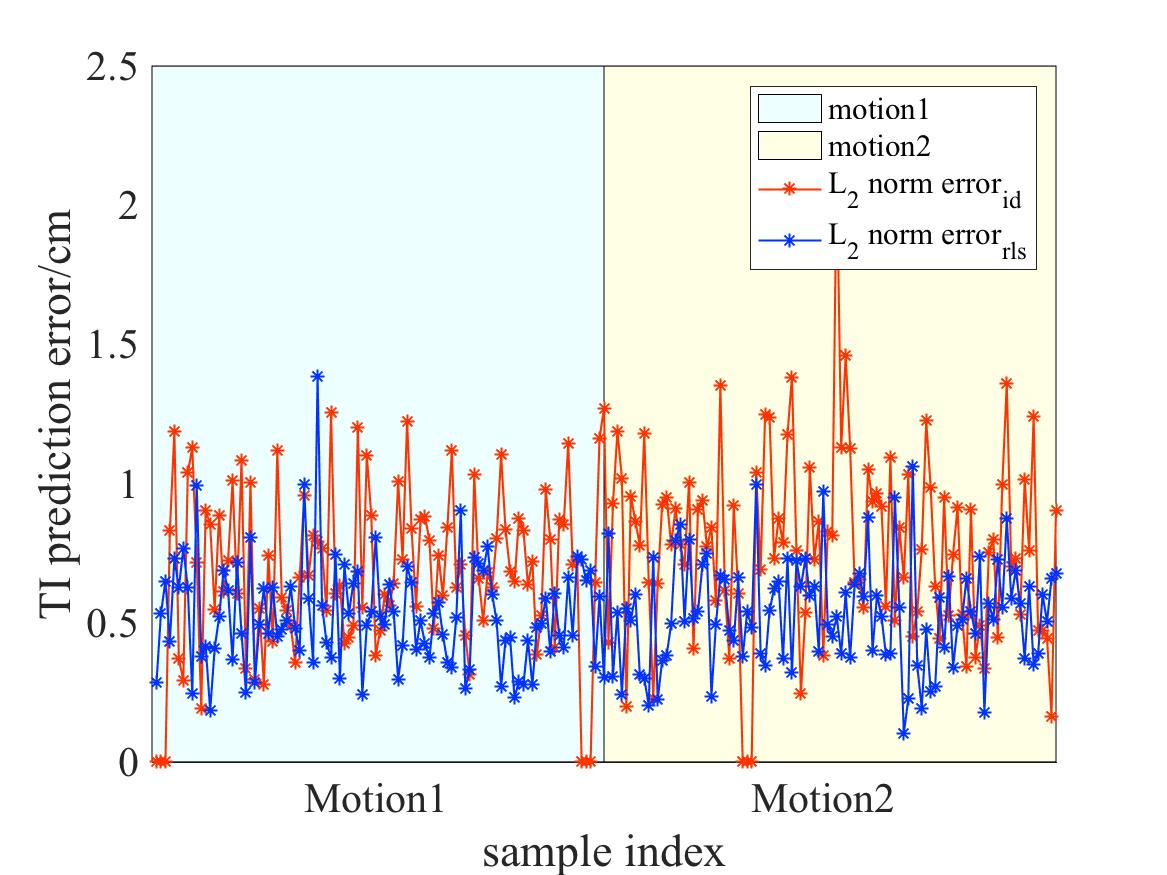}}
        \caption{Time Invariant dataset (k+1)}
    \end{subfigure}
    \hfill
    \begin{subfigure}[t]{0.32\textwidth}
        \raisebox{-\height}{\includegraphics[width=\textwidth]{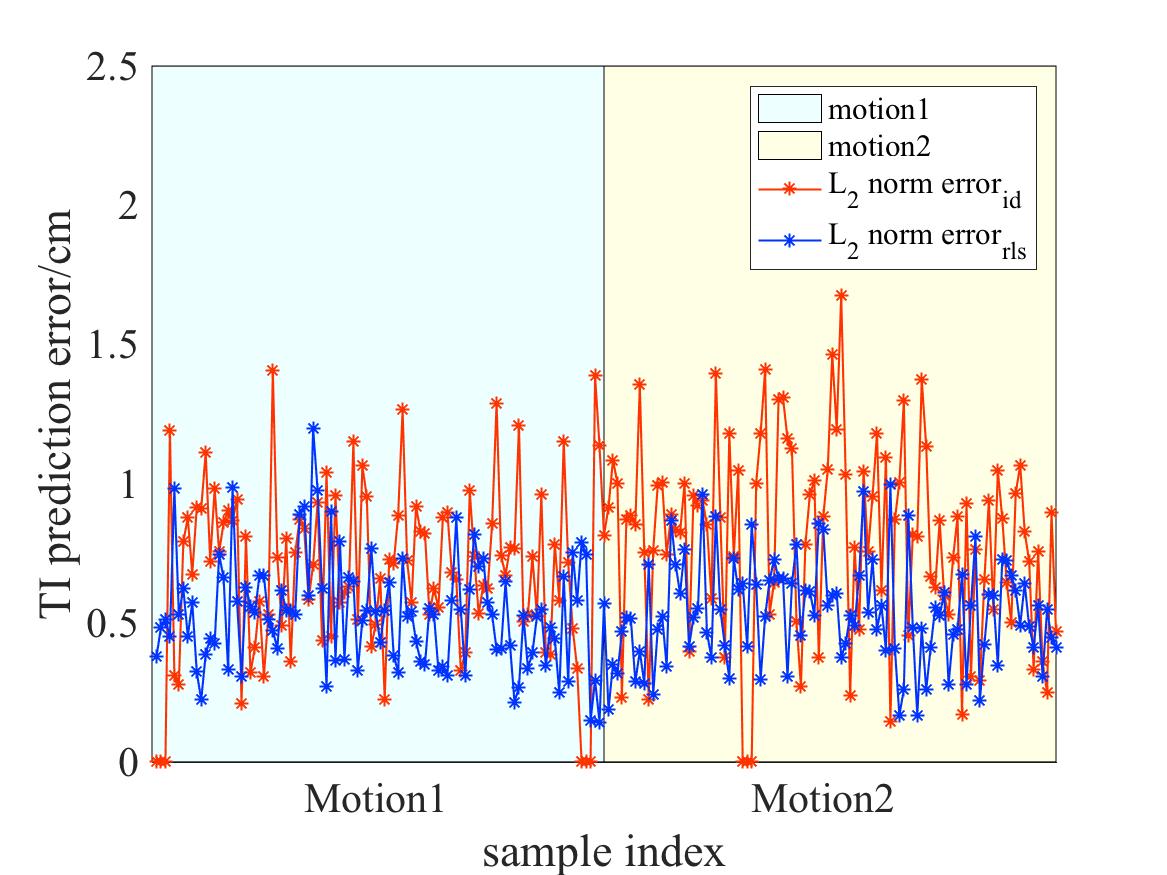}}
        \caption{Time Invariant dataset (k+2)}
    \end{subfigure}
    \hfill
    \begin{subfigure}[t]{0.32\textwidth}
        \raisebox{-\height}{\includegraphics[width=\textwidth]{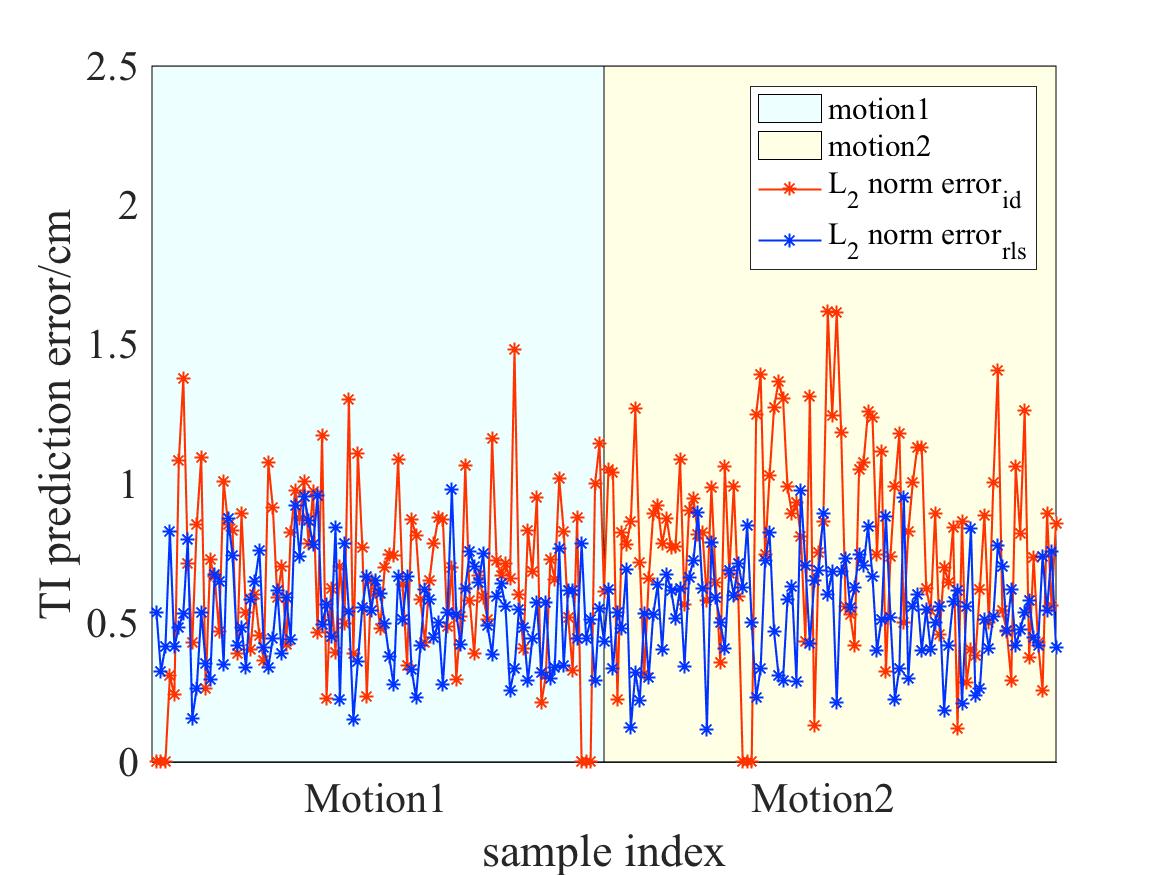}}
        \caption{Time Invariant dataset (k+3)}
    \end{subfigure}
    \begin{subfigure}[t]{0.32\textwidth}
        \raisebox{-\height}{\includegraphics[width=\textwidth]{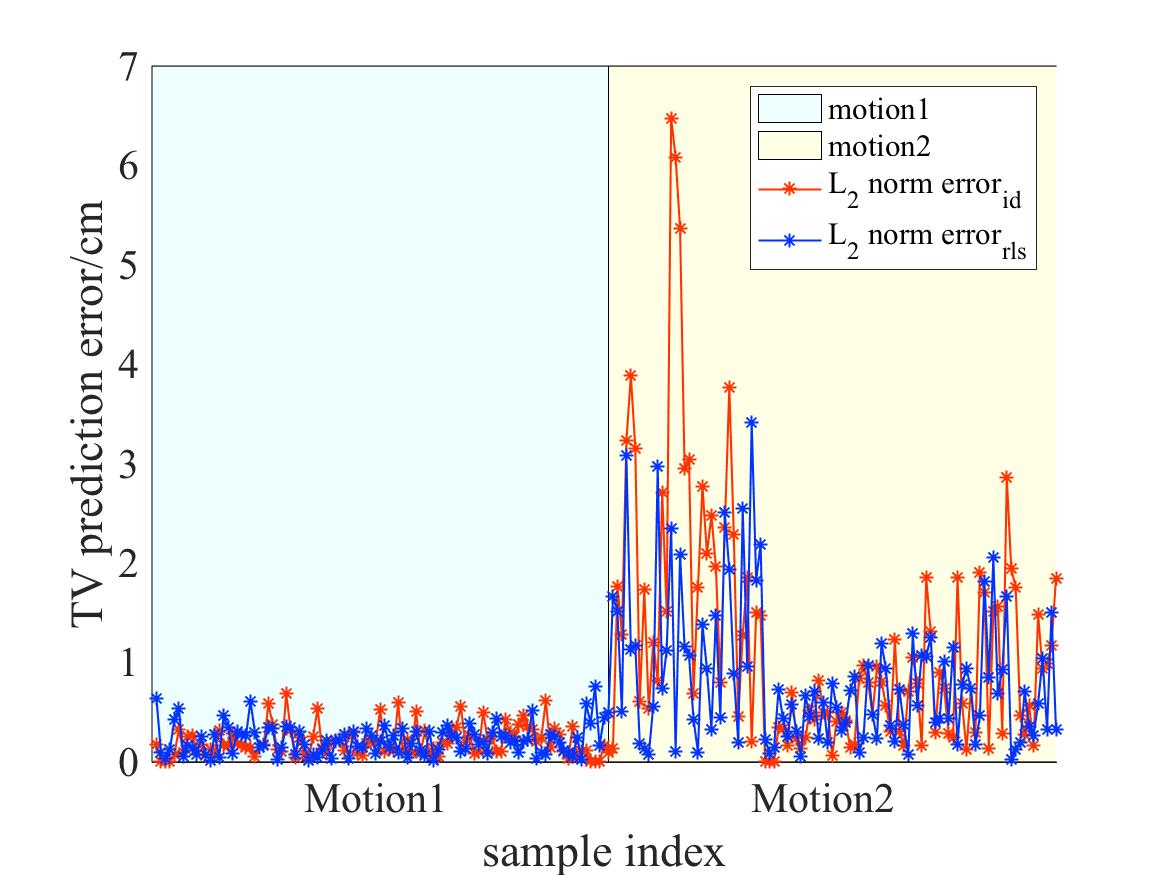}}
    \caption{Time Varying dataset (k+1)} 
    \end{subfigure}
    \hfill
    \begin{subfigure}[t]{0.32\textwidth}
        \raisebox{-\height}{\includegraphics[width=\textwidth]{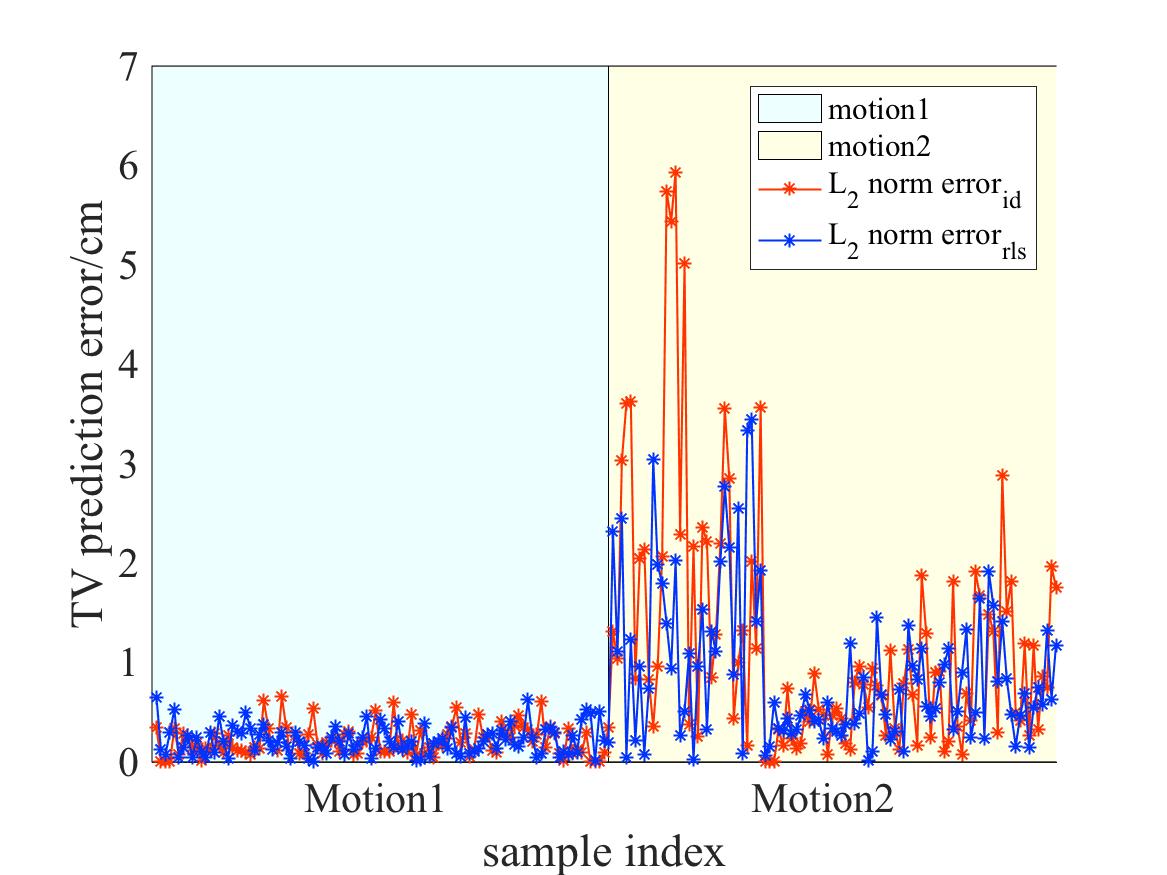}}
    \caption{Time Varying dataset (k+2)} 
    \end{subfigure}
    \hfill
    \begin{subfigure}[t]{0.32\textwidth}
        \raisebox{-\height}{\includegraphics[width=\textwidth]{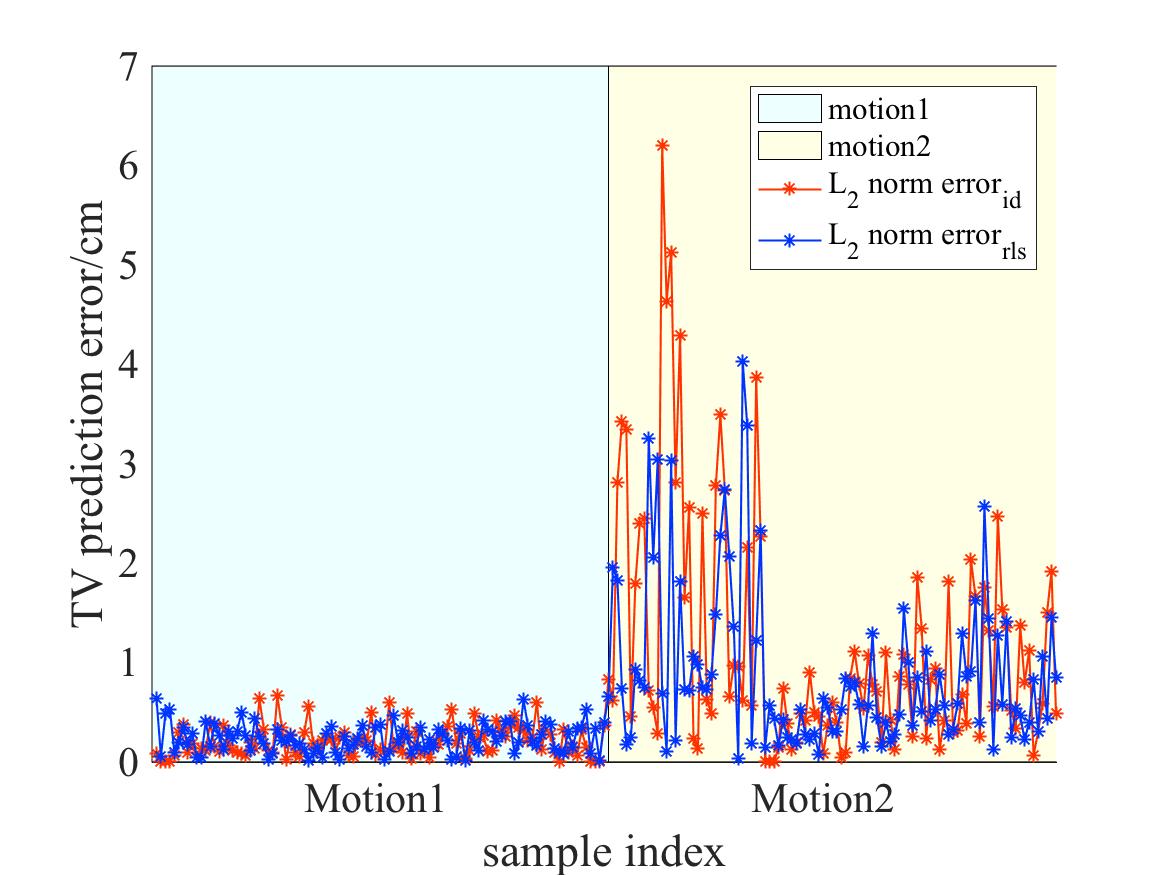}}
    \caption{Time Varying dataset (k+3)} 
    \end{subfigure}
    \begin{subfigure}[t]{0.32\textwidth}
        \raisebox{-\height}{\includegraphics[width=\textwidth]{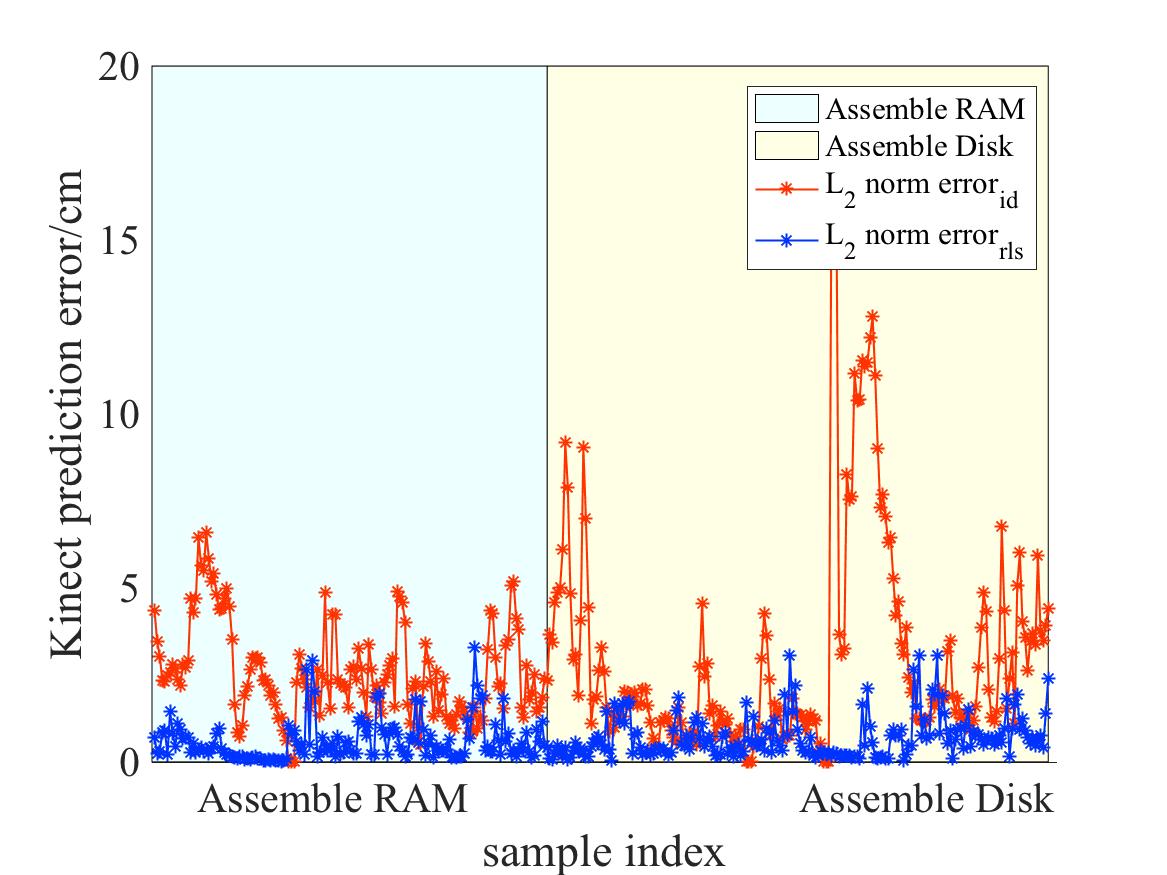}}
    \caption{Kinect dataset (k+1)} 
    \end{subfigure}
    \hfill
    \begin{subfigure}[t]{0.32\textwidth}
        \raisebox{-\height}{\includegraphics[width=\textwidth]{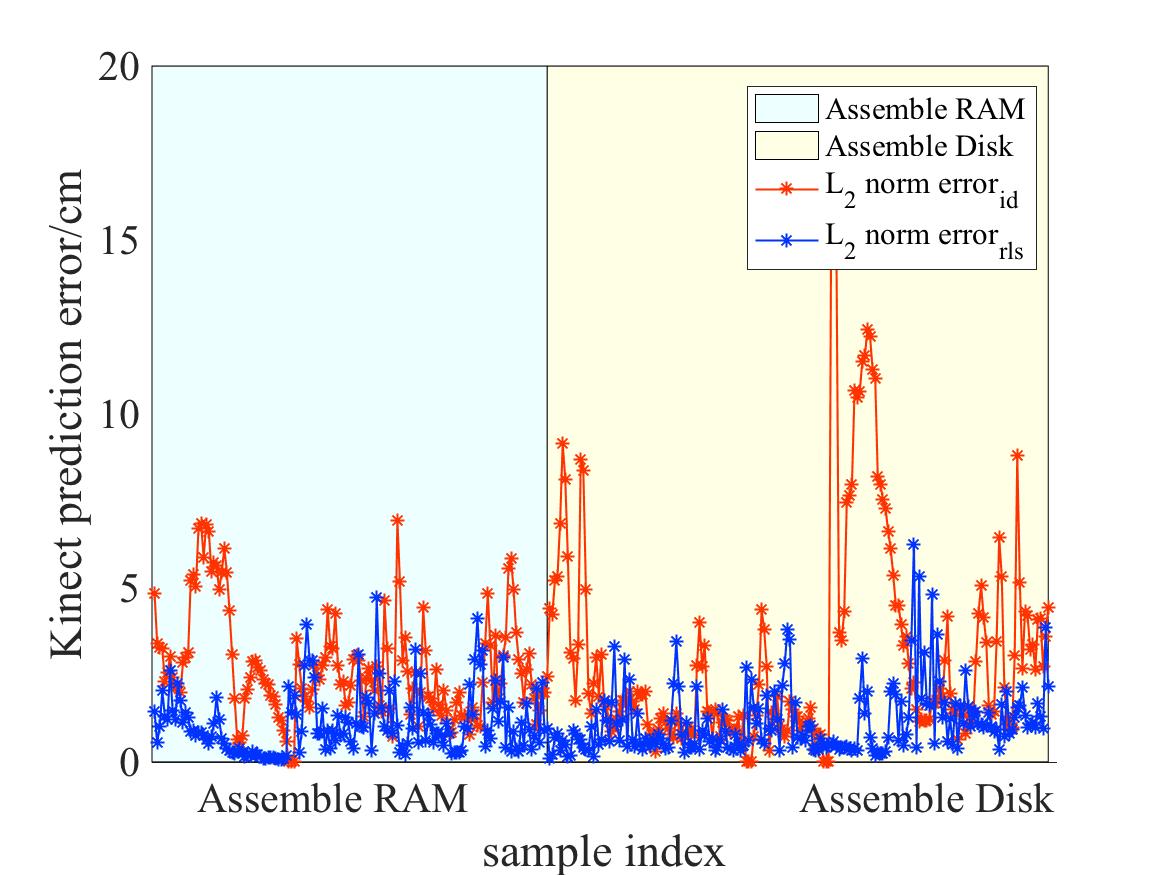}}
    \caption{Kinect dataset (k+2)} 
    \end{subfigure}
    \hfill
    \begin{subfigure}[t]{0.32\textwidth}
        \raisebox{-\height}{\includegraphics[width=\textwidth]{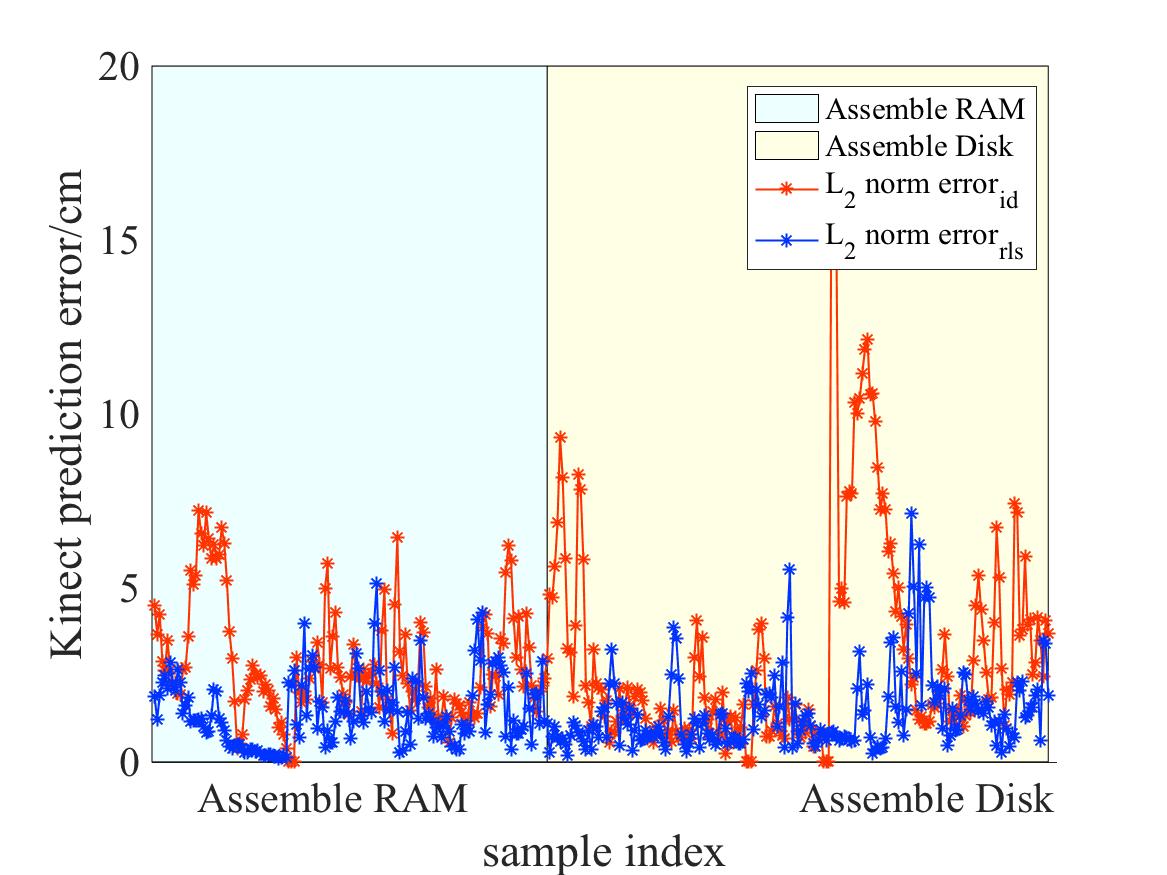}}
    \caption{Kinect dataset (k+3)} 
    \end{subfigure}
    \begin{subfigure}[t]{0.32\textwidth}
        \raisebox{-\height}{\includegraphics[width=\textwidth]{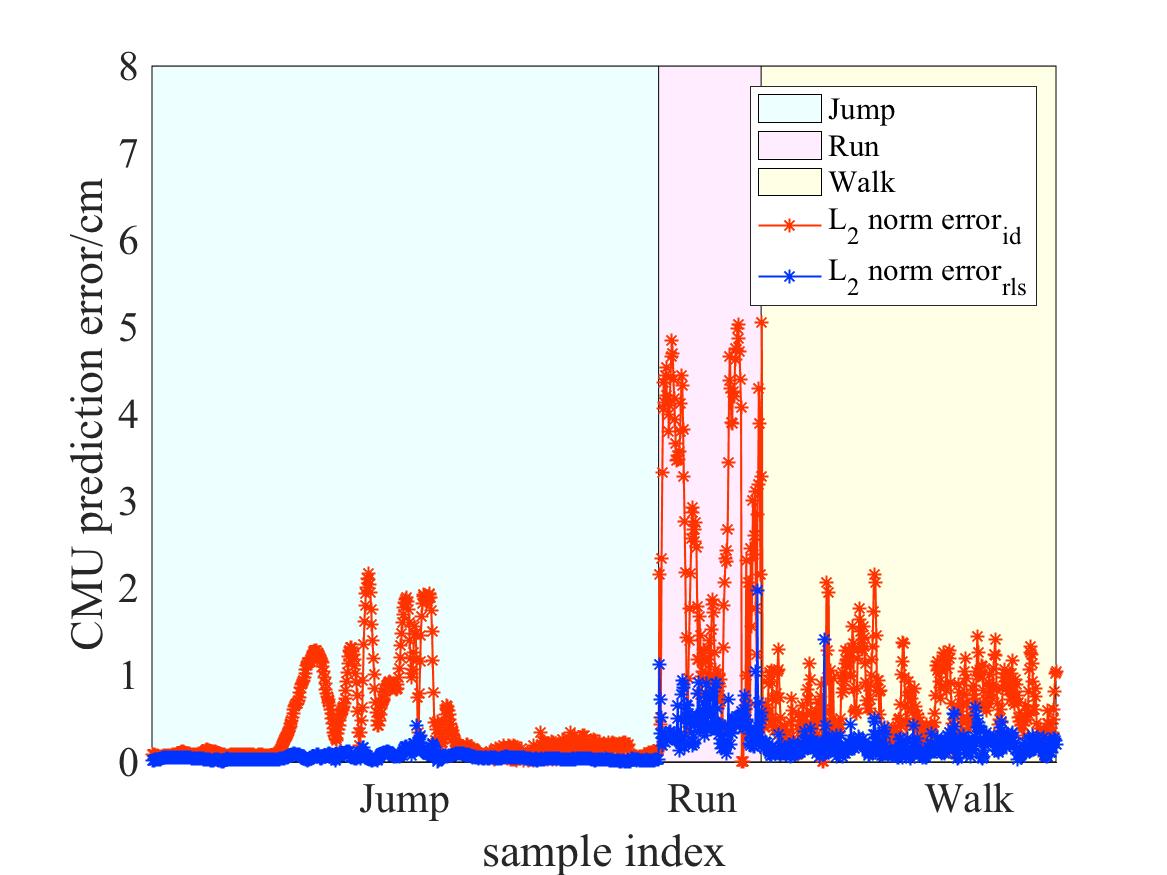}}
    \caption{CMU dataset (k+1)} 
    \end{subfigure}
    \hfill
    \begin{subfigure}[t]{0.32\textwidth}
        \raisebox{-\height}{\includegraphics[width=\textwidth]{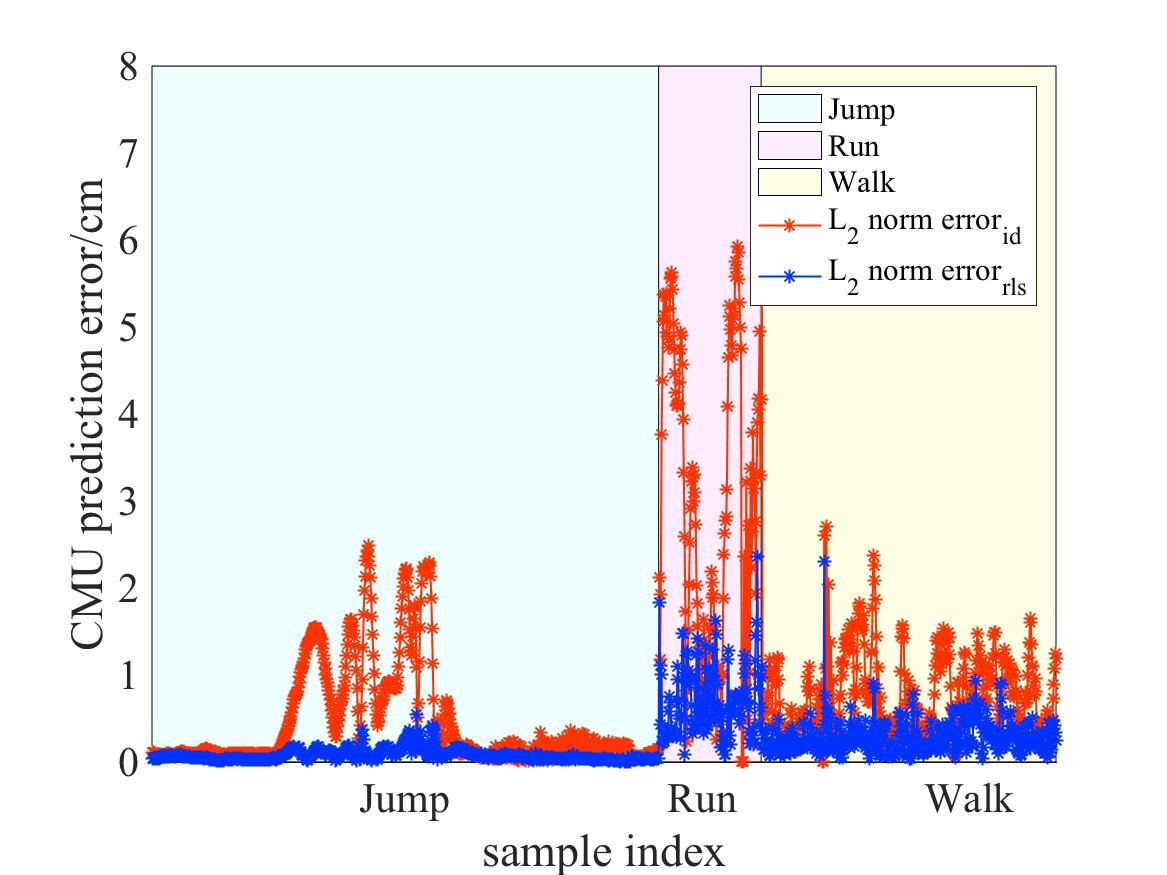}}
    \caption{CMU dataset (k+2)} 
    \end{subfigure}
    \hfill
    \begin{subfigure}[t]{0.32\textwidth}
        \raisebox{-\height}{\includegraphics[width=\textwidth]{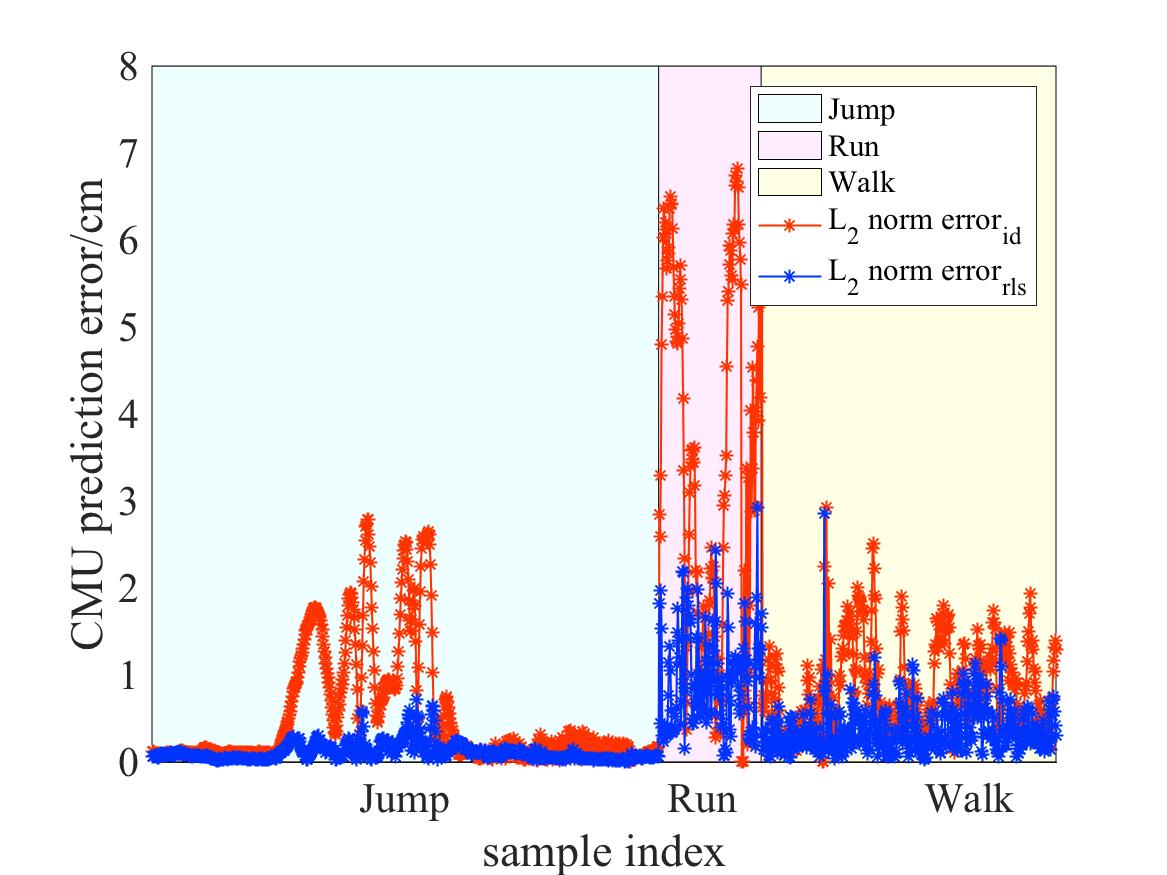}}
    \caption{CMU dataset (k+3)} 
    \end{subfigure}
    
     \caption{Prediction error comparison between RLS-PAA (blue system) and Identifier-based (red system) algorithm on four datasets. From top to bottom, each row is the experiment result tested on one trial of each motion on artificial time invariant system dataset, artificial time variant system dataset, Kinect dataset, and CMU dataset. In the experiment, when a new measurement is available, three future positions are predicted, and they are shown in the separate three columns respectively. The vertical black dash lines denote the boundaries of different motion classes.}
     \label{detailed}
\end{figure*}

We evaluate the performance of online adaption algorithms by prediction errors. The comparison between identifier-based algorithm and our methods is demonstrated in \cref{detailed}. We also compare the prediction errors by the neural network models without online adaptation.  RLS-PAA results in smallest prediction errors on all the datasets, which demonstrates the effectiveness of our proposed algorithm. Mean squared estimation error (MSEE) of all methods are shown in Table \ref{summarize}. Compared to identifier-based algorithm, RLS-PAA has much smaller MSEE across four different datasets in all $x, y, z$ axes. 

As for the training efficiency, since gradient operation is expensive to compute for large neural networks, we observed that it takes roughly $0.375$s for identifier-based algorithm to adapt one motion sample using standard gradient operation whereas it only takes $0.031$s  for our proposed method to adapt one sample. Our method is better for real-time applications. For identifier-based algorithm that adapts all parameters in the neural network, even though it is possible to accelerate gradient operation using central difference approximation, the computation complexity grows exponentially as the complexity of the neural network grows. All the experiments are performed on the MATLAB 2016 platform with $2.7 GHz$ Intel Core i5 Processor.

\begin{table}
 \centering
   \begin{tabular}{|p{2.5cm}|p{0.8cm}p{0.8cm}p{1cm}p{1cm}|}
   \hline
   & NN w ID & NN w RLS-PAA & NN w/o ID & NN w/o RLS-PAA\\
   \hline
   MSEE ${cm}^2$(CMU) & 21.9248 & \bf{5.5728} & 27.5121 & 30.2721 \\
   \hline
   \hline
   MSEE ${cm}^2$(Kinect) & 27.3493 & \bf{12.0419} & 29.6521 & 16.3973\\
   \hline
   \hline
   MSEE ${cm}^2$(TV) & 2.9093& \bf{2.2261} & 3.1889 & 2.5108 \\
   \hline
   \hline
   MSEE ${cm}^2$(TI) & 4.2925& \bf{1.3180} & 5.3221 & 4.2642\\
   \hline
   \hline
   \end{tabular}
 \caption{Mean Squared Estimation Error (MSEE) of predictions on four datasets for identifier-based algorithm (ID), recursive least square parameter adaptation algorithm (RLS-PAA), offline trained neural network in ID case without adaptation (NN w/o ID), and offline trained neural network in RLS-PAA case without adaptation (NN w/o RLS-PAA)}
 \label{summarize}
\end{table}

\section{Discussion\label{sec: discussion}}

The objective of this paper is to uncover a robust and general human motion transition model that can adapt to different motion patterns, and capture the uncertainty level of the prediction in real time. Our approach combines offline training and online adaptation for function approximation, which takes the advantages of both methods. Offline training of the neural network extracts global features from the data. Online adaptation changes the way that the global features are combined by adjusting the coefficients in the last layer of the network locally to minimize real time prediction error. Onerous model coefficient-tuning is not required in our approach. 

\subsection{Analyses} 
State-of-the-art identifier-based adaptation algorithm requires heavy coefficient tuning. The performance of identifier-based algorithm highly relies on the scale of update step size. The computation of the gradient has inertia that it cannot adapt the sudden change quickly. However, sudden change of velocity  happens a lot in human motion. Therefore, some peak prediction error can be observed from time to time as shown in \cref{detailed}, especially at the beginning of each motion trial, since there is a clear velocity jump between the beginning and the end of two different trials. 

The number of past joint positions being considered has impact on the prediction performance. RLS-PAA looks back into historical trajectory with exponentially decayed weights. The forgetting factor controls how far back the algorithm is using the past information. In our experiment on Kinect data, the optimal forgetting factor is $0.9998$. Note that to accommodate a more rapidly changing motion, the forgetting factor may be set smaller so that the algorithm looks back in a short range.

In many human robot interaction applications, human often work with robots over tables, standing or sitting in a fixed position where only the human upper body prediction is required. Note that human upper body can be viewed as a fixed-base cascade robot arm, which can be represented by a kinematic chain model. In our experiments, we focus on motion prediction for one wrist joint, which can be viewed as the end point position in the chain model. Whole arm movement can be predicted either using inverse kinematics or dynamic simulation given the prediction of the wrist position.

\subsection{Extensions}
Our proposed algorithm can be easily applied to many fields and extended by combination with other algorithms. Generally speaking, we can apply the method to many other time series data prediction, such as vehicle and pedestrian motion prediction. In these cases, we regard the vehicles and pedestrians as joints in our method, and adapt the model and obtain uncertainty level online, which is beneficial in autonomous driving. Our proposed algorithm can also be combined with other algorithms by modeling the velocity state transition pattern, which can be used to adjust the update step size.

\section{Conclusion\label{sec: conclusion}}
This paper proposed a semi-adaptable neural network to predict human motion, which is capable of adapting the model online and provides uncertainty level for safety constraints. Offline trained neural network takes the historical joint trajectory as input and outputs the predictions. In online test, the parameters of the last layer in the neural network model are adapted to accommodate individual differences and time-varying behaviors. The extensive experiment results demonstrate that our method significantly outperforms the state-of-the-art method on the majority of the motion datasets in terms of prediction error. Moreover, our model is much more computationally efficient and is free of onerous tuning. Moreover, the performance is robust across different motion categories and datasets. 

\bibliography{main}
\bibliographystyle{IEEEtran}

\end{document}